\documentclass[]{fairmeta}
\usepackage{microtype}
\usepackage{lmodern}
\usepackage{amsfonts}
\usepackage{graphicx}
\usepackage{booktabs}
\usepackage{float}
\usepackage{wrapfig}
\usepackage{amsmath}
\usepackage{amssymb}
\usepackage[most]{tcolorbox}
\usepackage{amsthm}
\usepackage{algorithm}
\usepackage{algorithmic}
\usepackage[capitalize,noabbrev]{cleveref}
\theoremstyle{plain}

\theoremstyle{definition}

\theoremstyle{remark}

\usepackage{enumitem}
\usepackage{xcolor}
\usepackage{url}
\usepackage{hyperref}
\usepackage{nicefrac}

\title{%
  \raisebox{-0.3\height}{\includegraphics[width=0.3\textwidth]{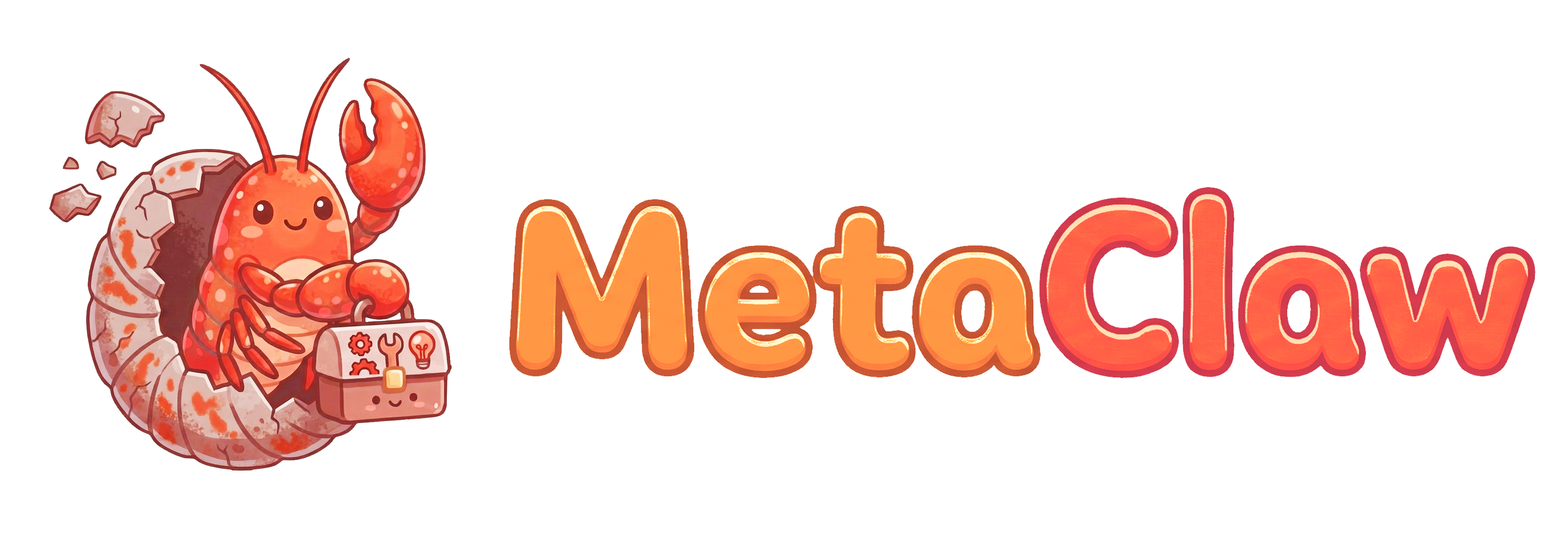}}%
  : Just Talk -- An Agent That Meta-Learns and Evolves in the Wild%
}

\author[1*]{Peng Xia}
\author[1*]{Jianwen Chen}
\author[2*]{Xinyu Yang}
\author[3*]{Haoqin Tu}
\author[1*]{Jiaqi Liu}
\author[1*]{Kaiwen Xiong}
\author[1]{Siwei Han}
\author[1]{Shi Qiu}
\author[1]{Haonian Ji}
\author[3]{Yuyin Zhou}
\author[4]{Zeyu Zheng}
\author[3]{Cihang Xie}
\author[1*]{Huaxiu Yao}

\affiliation[1]{{\includegraphics[width=0.02\textwidth]{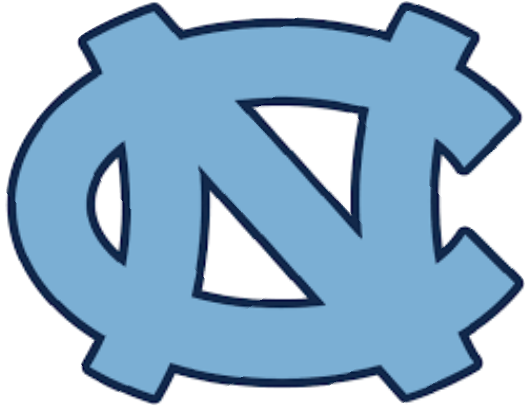}} UNC-Chapel Hill}
\affiliation[2]{{\includegraphics[width=0.015\textwidth]{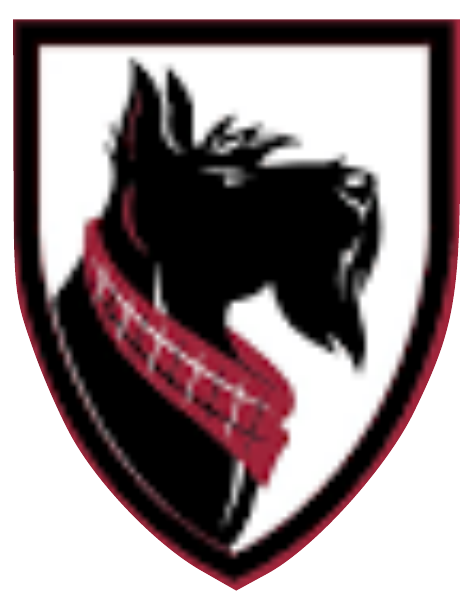}} Carnegie Mellon University}
\affiliation[3]{{\includegraphics[width=0.02\textwidth]{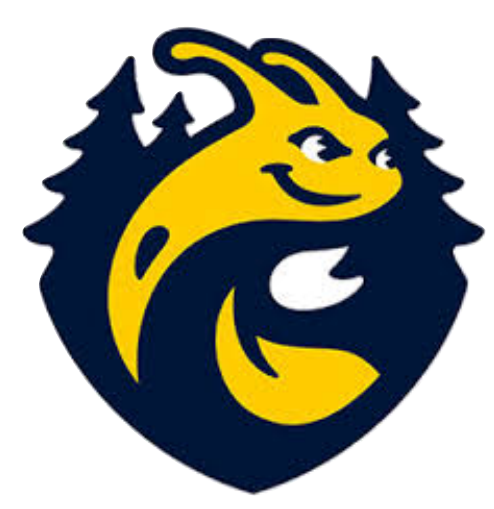}} UC Santa Cruz}
\affiliation[4]{{\includegraphics[width=0.02\textwidth]{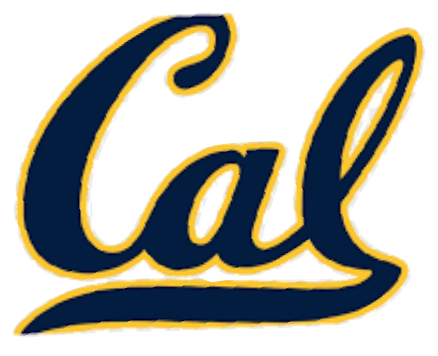}} UC Berkeley}
\affiliation[*]{Core Contributors}

\abstract{
Large language model (LLM) agents have rapidly emerged as powerful assistants for complex, multi-step tasks, yet agents deployed in the wild remain largely static, trained once and served unchanged regardless of how user needs evolve. This creates a fundamental tension: they must serve users continuously without interruption, yet their capabilities grow stale as the task distribution drifts with real-world usage. On platforms such as OpenClaw, where a single agent connects to 20+ messaging channels and handles diverse, evolving workloads, existing approaches either store raw trajectories without distilling transferable behavioral knowledge, maintain static skill libraries disconnected from weight optimization, or incur service downtime during retraining. We present \textbf{MetaClaw}, a continual meta-learning framework that jointly maintains a base LLM policy and an evolving skill library of reusable behavioral instructions, improving both through two complementary mechanisms. \emph{Skill-driven fast adaptation} analyzes failure trajectories and synthesizes new skills via an LLM evolver, taking effect immediately with zero service downtime. \emph{Opportunistic policy optimization} performs gradient-based weight updates via cloud LoRA fine-tuning using RL with a process reward model, triggered only during user-inactive windows by the Opportunistic Meta-Learning Scheduler (OMLS), which monitors configurable sleep hours, system keyboard inactivity, and Google Calendar occupancy. The two mechanisms are mutually reinforcing: a better policy produces more informative failures for skill synthesis, and richer skills yield higher-reward trajectories for policy optimization. To prevent stale reward contamination, a skill generation versioning mechanism strictly separates support data (failure trajectories consumed by skill evolution) from query data (post-adaptation trajectories used for RL updates). Built on a proxy-based architecture, MetaClaw scales to production-size LLMs without a local GPU. Experiments on MetaClaw-Bench (934 questions, 44 simulated workdays) and AutoResearchClaw (23-stage autonomous research pipeline) demonstrate consistent improvements: skill-driven adaptation improves accuracy by up to 32\% relative; the full pipeline advances Kimi-K2.5 from 21.4\% to 40.6\% accuracy (vs.\ GPT-5.2 baseline 41.1\%) with an 8.25$\times$ gain in end-to-end task completion; and skill injection alone improves AutoResearchClaw composite robustness by 18.3\%.
}

\metadata[Github]{\url{https://github.com/aiming-lab/MetaClaw}}

\begin{document}

\maketitle

\section{Introduction}

\begin{figure}[t]
    \centering
    \includegraphics[width=0.95\linewidth]{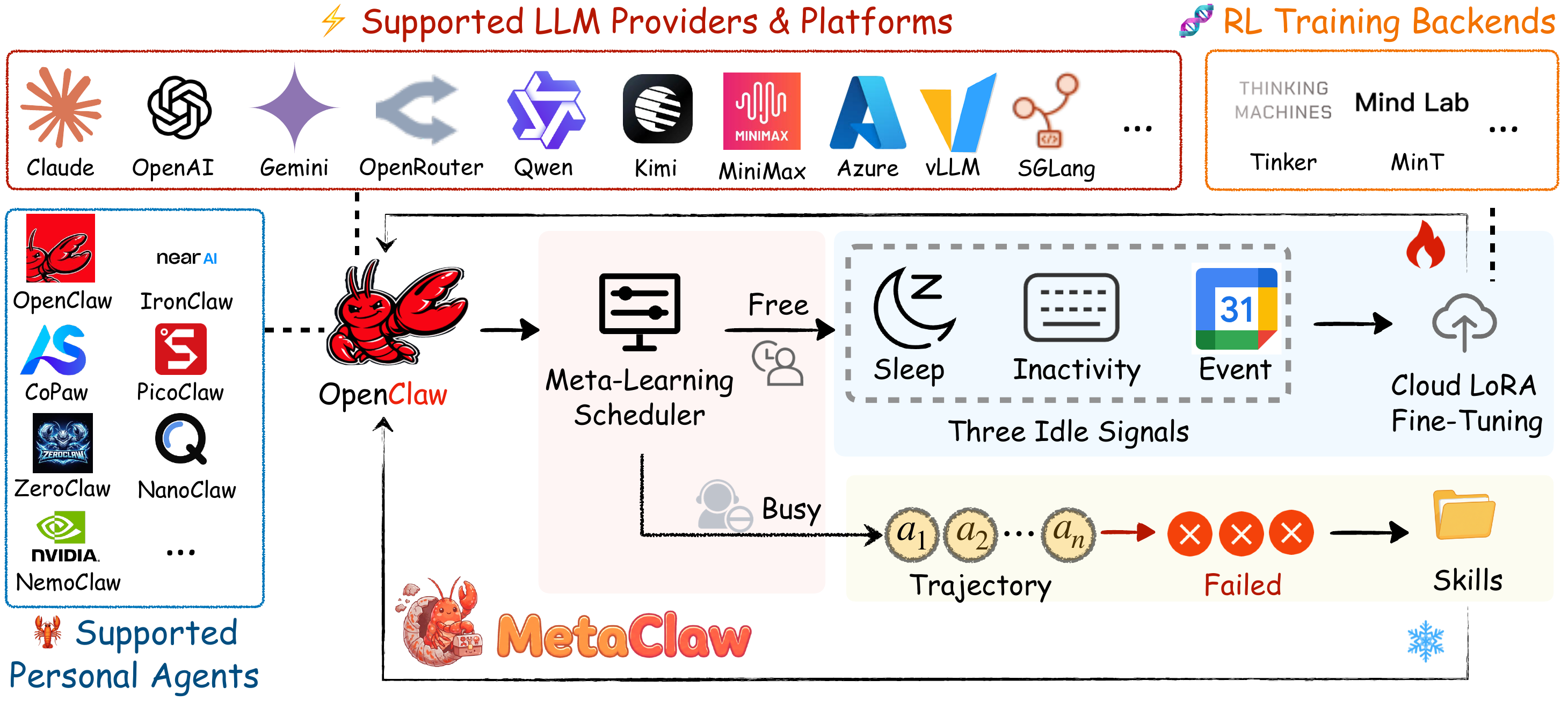}
    \caption[Overview of MetaClaw.]{Overview of \raisebox{-0.25\height}{\includegraphics[height=1.8em]{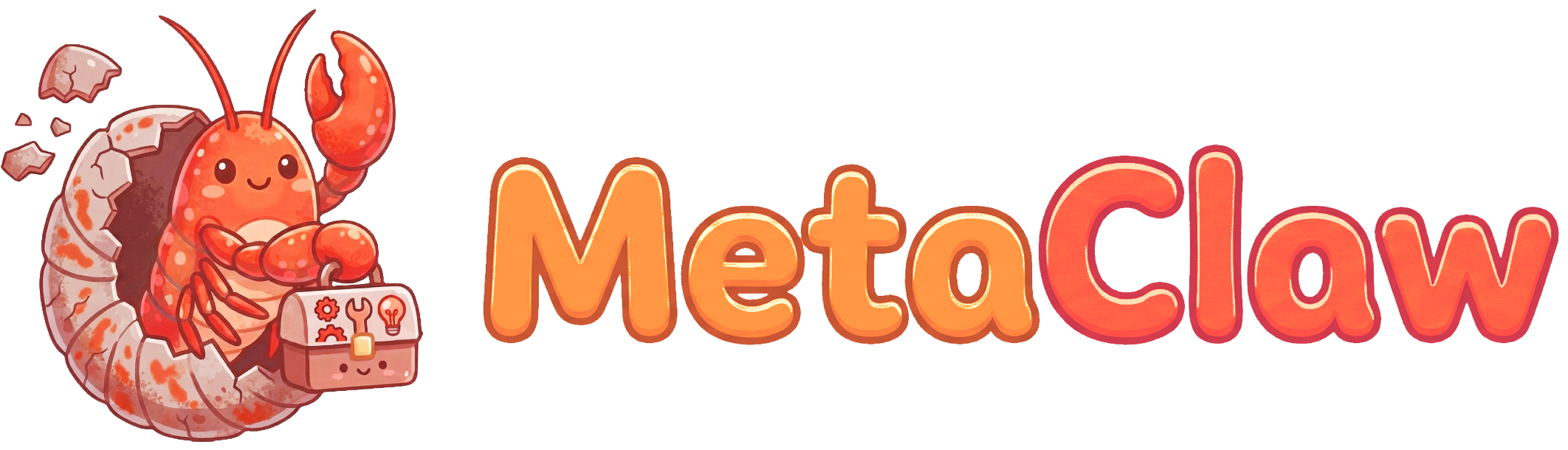}}. The framework improves the meta-model $\mathcal{M} = (\theta, \mathcal{S})$ via two complementary loops operating at different timescales. Skill-driven fast adaptation (left) analyzes failed trajectories and instantly expands the skill library $\mathcal{S}$ without parameter updates, taking effect immediately for subsequent tasks. Opportunistic policy optimization (right) accumulates post-adaptation trajectories and, once sufficient data is available, leverages idle signals (sleep, inactivity, calendar) detected by the Opportunistic Meta-Learning Scheduler to trigger RL-based weight updates on $\theta$ via Cloud LoRA fine-tuning.}
    \label{fig:overview}
    \vspace{-1.5em}
\end{figure}

Large language model (LLM) agents have demonstrated remarkable capabilities across complex tasks~\citep{yao2022react,shinn2023reflexion}, yet agents deployed in the wild remain largely static, trained once and served unchanged regardless of how the user's needs evolve~\citep{zhang2025agentracer,naihin2023testing,song2026agents}. Consider OpenClaw~\citep{openclaw}, an open-source CLI agent platform connecting to 20+ messaging channels, where a single user's workload may shift from multi-step file system operations one week to multi-agent messaging workflows the next. As the task distribution drifts, a frozen model becomes increasingly misaligned with actual usage patterns, repeatedly failing on task types underrepresented during pretraining.

Existing approaches to agent adaptation fall into three broad categories, each with notable limitations. Memory-based methods~\citep{shinn2023reflexion,zhao2024expel,fang2025memp,tang2025agent,ouyang2025reasoningbank,chhikara2025mem0,liu2026simplemem} store raw conversation trajectories for future retrieval, but such trajectories are verbose and redundant, preventing the agent from extracting transferable behavioral patterns. Skill-based methods~\citep{xia2026skillrl,zhang2025memevolve,zhang2026memrl,wu2025evolver,zhang2026memskill} compress experience into reusable behavioral instructions, yet treat the resulting skill library as a static database never coordinated with weight optimization. RL-based methods~\citep{schulman2017proximal,ahmadian2024back,shao2024deepseekmath,feng2025group,zheng2025group} update model weights, but operate in small-scale or offline settings and ignore a critical data validity problem: once skills have evolved, trajectories collected under the old skill context carry stale rewards that contaminate gradient updates if reused without filtration. A common thread across all three categories is that each addresses only one aspect of adaptation in isolation, leaving the complementary dimensions unexploited.

Our key observation is that two fundamentally different timescales of adaptation are in fact naturally complementary. Behavioral heuristics (e.g., ``always verify a file path before reading,'' ``confirm before destructive commands'') can be distilled within seconds from a single failed conversation and injected immediately as skill instructions. Improving the model's underlying policy across diverse task types requires gradient-based optimization over many trajectories, on a timescale of minutes to hours. The two mechanisms are also mutually reinforcing: a better policy produces more informative failures for skill synthesis, and richer skills yield higher-reward trajectories for policy optimization. No existing system unifies these two forms of adaptation into a coherent framework that exploits this virtuous cycle.

We present \textbf{MetaClaw}, a continual meta-learning~\citep{finn2017model,yao2021meta} framework that jointly maintains a base LLM policy and an evolving skill library of reusable behavioral instructions. The skill library serves a dual role: as a \emph{meta-parameter} that accumulates behavioral knowledge across the task stream, and as an \emph{adaptation basis} from which task-specific skills are retrieved at inference time. MetaClaw improves both components through two mechanisms. \emph{Skill-driven fast adaptation} performs gradient-free skill evolution: an LLM analyzes failure trajectories and synthesizes new behavioral instructions~\citep{xia2026skillrl} that take effect immediately with zero service downtime. \emph{Opportunistic policy optimization} uses RL with a process reward model (PRM)~\citep{zhang2025lessons} to update model weights via cloud~\citep{tinker} LoRA fine-tuning~\citep{hu2021lora}, optimizing post-adaptation performance. Two design principles govern their coordination. First, \emph{when} to run policy optimization: our Opportunistic Meta-Learning Scheduler (OMLS) monitors three idle signals, i.e., configurable sleep hours, system keyboard inactivity, and Google Calendar event occupancy, and triggers weight updates only during user-inactive windows, eliminating downtime. Second, \emph{which data} to use: we distinguish \emph{support data} (failure trajectories consumed by skill evolution) from \emph{query data} (trajectories collected after new skills take effect). Only query data, reflecting the agent's post-adaptation behavior, is valid for RL; support data carries rewards conditioned on the old skill context and is excluded. Our skill generation versioning mechanism enforces this separation by stamping each trajectory with its skill generation index and flushing stale samples from the training buffer whenever skills evolve.

In summary, our primary contribution is MetaClaw, a continual meta-learning framework that unifies skill-driven fast adaptation with opportunistic policy optimization, enabling deployed LLM agents to evolve continuously through a proxy-based architecture without requiring a local GPU. We evaluate on MetaClaw-Bench, a new benchmark of 934 questions over 44 simulated workdays, where each day forms a sequential, feedback-driven multi-round session of real CLI tasks (file editing, JSON structuring, shell scripting). Experiments with GPT-5.2 and Kimi-K2.5 show that skill-driven fast adaptation alone improves overall accuracy by up to 32.2\% in relative terms; MetaClaw (Full) further advances Kimi-K2.5 from 21.4\% to 40.6\%, improves end-to-end task completion by 8.25$\times$ on Part~I and file-check completion by 185\% on Part~II, and nearly closes the gap with GPT-5.2's baseline. We further validate on AutoResearchClaw, a 23-stage autonomous research pipeline, where skill injection alone improves the composite robustness score by 18.3\%, demonstrating cross-domain generalization of MetaClaw's adaptation mechanisms.

\section{Problem Setup}
\label{sec:setup}

We consider a deployed CLI agent that serves a user over a stream of tasks $\tau_1, \tau_2, \ldots$ drawn from a non-stationary distribution $p_t(\tau)$. Each task $\tau_i$ consists of a user instruction and environmental context (file system state, shell history, etc.), and the agent must produce a sequence of actions $a_{1:T}$ to accomplish the task. The agent's behavior at any point in time is fully determined by a \emph{meta-model}:
\begin{equation}
    \mathcal{M} = (\theta, \mathcal{S}),
\end{equation}
where $\theta$ denotes the parameters of the base LLM policy and $\mathcal{S} = \{s_1, s_2, \ldots, s_K\}$ is a library of \emph{skill instructions}, i.e., concise, reusable behavioral directives injected into the agent's system prompt at inference time. Given a task $\tau$, the agent generates actions according to:
\begin{equation}
    a \sim \pi_\theta\!\left(\cdot \mid \tau,\; \textsc{Retrieve}(\mathcal{S}, \tau)\right),
\end{equation}
where $\textsc{Retrieve}(\mathcal{S}, \tau) \subseteq \mathcal{S}$ selects the most relevant skills for the current task via embedding-based retrieval.

The meta-model $\mathcal{M}$ evolves over the task stream as the agent accumulates experience. We distinguish two types of trajectory data based on their role in this evolution. \emph{Support data} $\mathcal{D}^{\text{sup}}$ consists of trajectories whose failures drive adaptation of the skill library $\mathcal{S}$; these trajectories are consumed by the adaptation process and reflect pre-adaptation behavior. \emph{Query data} $\mathcal{D}^{\text{qry}}$ consists of trajectories collected after adaptation has taken effect; these reflect the agent's post-adaptation behavior and are used to optimize the policy parameters $\theta$. Maintaining a strict separation between support and query data is essential: mixing them would cause $\theta$ to be optimized against stale reward signals that no longer reflect the agent's current capabilities.

The goal of MetaClaw is to continuously improve $\mathcal{M}$ over the task stream, not merely to solve each task in isolation, but to \emph{become progressively better at adapting} to new tasks as they arrive. This positions MetaClaw as a continual meta-learning system: the agent learns from a non-stationary task stream while simultaneously improving its own adaptation capability.

\section{MetaClaw}
\label{sec:method}

\subsection{Overview}

MetaClaw improves the meta-model $\mathcal{M} = (\theta, \mathcal{S})$ through two complementary mechanisms operating at different timescales (Figure~\ref{fig:overview}). \emph{Skill-driven fast adaptation} analyzes failure trajectories and synthesizes new skill instructions that are immediately injected into the agent's prompt, evolving $\mathcal{S}$ without touching model weights. \emph{Opportunistic policy optimization} uses post-adaptation trajectories to update $\theta$ via reinforcement learning, deferred to user-inactive windows by the Opportunistic Meta-Learning Scheduler (OMLS). A skill generation versioning mechanism ensures that policy optimization always trains on query data collected under the current skill library, preventing stale reward contamination from support data. The two mechanisms are mutually reinforcing: a better $\theta$ produces more informative failures for skill synthesis, and richer skills produce higher-reward trajectories for policy optimization. This virtuous cycle enables the system to \emph{learn to become better at adapting}. The complete procedure is summarized in Algorithm~\ref{alg:metaclaw}.

\subsection{Skill-Driven Fast Adaptation}

Given the current meta-model $(\theta, \mathcal{S}_g)$, the agent executes tasks and collects trajectories. Trajectories that reveal failure modes form the support set $\mathcal{D}^{\text{sup}}_g$. Skill-driven adaptation evolves the skill library via a gradient-free experience distillation process:
\begin{equation}
    \mathcal{S}_{g+1} = \mathcal{S}_g \cup \mathcal{E}(\mathcal{S}_g, \mathcal{D}^{\text{sup}}_g),
\end{equation}
where $\mathcal{E}$ is a \emph{skill evolver}, an LLM that analyzes failure trajectories and synthesizes new behavioral instructions. The index $g$ denotes the \emph{skill generation}, incremented each time the library changes. This step modifies only $\mathcal{S}$, leaving $\theta$ fixed, and takes effect immediately for all subsequent tasks. Because skill injection operates through the prompt rather than model parameters, fast adaptation incurs zero service downtime.

This mechanism is gradient-free by design, not by approximation. The skill library $\mathcal{S}$ lives in a discrete natural-language space where gradient descent is ill-defined; LLM-based failure analysis is the natural adaptation mechanism for this space.

The skill library $\mathcal{S}$ plays a dual role in the learning structure. As a \emph{meta-parameter}, $\mathcal{S}$ accumulates behavioral knowledge across the entire task stream, with each skill generation $\mathcal{S}_{g+1} \supseteq \mathcal{S}_g$ representing the system's growing operational knowledge. As an \emph{adaptation basis}, $\textsc{Retrieve}(\mathcal{S}, \tau)$ extracts a task-specific subset at inference time, providing instant specialization without any parameter update. This dual character arises because natural-language instructions are inherently cross-task transferable: a skill distilled from one failure (e.g., ``verify file path before reading'') generalizes to all tasks involving file operations. Unlike systems where task-specific adaptations are ephemeral and discarded after each task, each adaptation episode in MetaClaw contributes lasting knowledge to the meta-model, making knowledge accumulation a feature rather than a side effect.

\subsection{Opportunistic Policy Optimization}

After each skill-driven adaptation step, the agent continues serving tasks under the latest skill library. Because policy optimization is deferred to idle windows, the skill library may have advanced through several generations by the time training begins. Let $g^*$ denote the current skill generation when a training window opens. The RL buffer $\mathcal{B}$ accumulates query trajectories across all post-adaptation generations, and policy optimization updates $\theta$ over this buffer:
\begin{equation}
    \theta_{t+1} = \theta_t + \alpha \nabla_\theta \mathbb{E}_{(\tau, \xi, g') \sim \mathcal{B}}\!\left[R(\pi_\theta(\cdot \mid \tau, \mathcal{S}_{g'}))\right],
    \label{eq:outer}
\end{equation}
where $g' \leq g^*$ is the skill generation under which each trajectory was collected, and $R$ is a process reward model (PRM) score. The versioning mechanism (Section~\ref{sec:versioning}) guarantees that $\mathcal{B}$ contains only query data, i.e., every sample reflects post-adaptation behavior under its respective skill generation. Crucially, policy optimization does not optimize $\theta$ for raw task performance, but for how well the agent performs \emph{after skill adaptation}. A better $\theta$ yields a meta-model from which skill-driven adaptation produces stronger post-adaptation behavior, resulting in an improved meta-model $\mathcal{M}' = (\theta_{t+1}, \mathcal{S}_{g^*})$.

In practice, policy optimization is realized via cloud LoRA fine-tuning using GRPO, deferred to idle windows by the Opportunistic Meta-Learning Scheduler (Section~\ref{sec:scheduler}). Importantly, training is initiated only after the query buffer $\mathcal{B}$ has accumulated a sufficient number of trajectories; launching RL with too few samples leads to high-variance gradient estimates and unstable policy updates. This means policy optimization naturally lags behind skill-driven adaptation by days or longer, reinforcing the asymmetry between the two timescales: skills evolve continuously, while the policy improves in discrete, data-gated steps.

\subsection{Skill Generation Versioning}
\label{sec:versioning}

The support-query separation defined in Section~\ref{sec:setup} must be enforced in MetaClaw's online setting, where tasks arrive sequentially and skill evolution is triggered asynchronously. Without a dedicated mechanism, support data can leak into the policy optimization buffer.

The problem is concrete: a trajectory $(\tau_i, \xi_i)$ that triggers skill evolution from $\mathcal{S}_g$ to $\mathcal{S}_{g+1}$ carries a reward $r_i$ reflecting performance under $\mathcal{S}_g$, \emph{before} the new skill existed. If this trajectory enters the RL buffer, policy optimization receives a gradient that penalizes $\theta$ for a failure that skill-driven adaptation has already corrected, optimizing for pre-adaptation rather than post-adaptation performance and violating the meta-learning objective in Eq.~\ref{eq:outer}.

We enforce separation via a \emph{skill generation version} $g_i$ stamped on each collected sample:
\begin{itemize}[leftmargin=*]
    \item \textbf{Support set} $\mathcal{D}^{\text{sup}}_g$: trajectories collected under $\mathcal{S}_g$ whose failures trigger skill evolution $\mathcal{S}_g \to \mathcal{S}_{g+1}$. These are consumed by the skill evolver and \emph{discarded from the RL buffer}.
    \item \textbf{Query set} $\mathcal{D}^{\text{qry}}_{g+1}$: trajectories collected after $\mathcal{S}_{g+1}$ takes effect. Only these, reflecting the agent's post-adaptation behavior, are eligible for policy optimization gradient updates.
\end{itemize}

When the skill generation counter advances from $g$ to $g+1$, the trainer flushes all samples with version $\leq g$ from its buffer. This ensures policy optimization always updates $\theta$ with respect to the agent's adapted behavior, preserving the integrity of the meta-learning structure.

\subsection{Opportunistic Meta-Learning Scheduler}
\label{sec:scheduler}

Policy optimization requires a model weight hot-swap upon completion, which briefly interrupts inference. In a deployed interactive system, this creates a tension: policy optimization must run periodically to improve $\theta$, but it must not degrade the user's experience.

We introduce the \emph{Opportunistic Meta-Learning Scheduler} (OMLS), a background daemon that defers policy optimization to periods when the user is not actively interacting with the agent. OMLS monitors three complementary idle signals:

\noindent \textbf{(1) Sleep window.} The user configures a sleep schedule (e.g., 23:00--07:00). During this window, the system is guaranteed to be idle, providing the largest contiguous training block.

\noindent \textbf{(2) System inactivity.} OMLS polls the operating system's input device idle timer (e.g., \texttt{ioreg HIDIdleTime} on macOS). If no keyboard or mouse activity is detected for $\delta$ minutes (default: 30), a training window opens. Upon renewed input, the trainer pauses gracefully via mid-batch checkpointing.

\noindent \textbf{(3) Calendar-aware scheduling.} OMLS queries the user's Google Calendar API. When the current time falls within a scheduled meeting, the user is presumed unavailable, opening an opportunistic training window. This is the most anticipatory of the three signals: it leverages the user's own schedule to predict idle periods proactively.

A training window opens when \emph{any} signal indicates user absence and closes when \emph{any} signal indicates the user has returned. The RL trainer supports pause/resume across fragmented idle windows, accumulating gradient steps opportunistically without requiring a single long contiguous block.

\begin{algorithm}[t]
\caption{MetaClaw: Continual Meta-Learning for Deployed LLM Agents}
\label{alg:metaclaw}
\small
\begin{algorithmic}[1]
\REQUIRE Meta-model $\mathcal{M} = (\theta_0, \mathcal{S}_0)$, skill evolver $\mathcal{E}$, task stream $\{\tau_i\}$, PRM $R$, OMLS idle detector
\ENSURE Continuously improved meta-model $\mathcal{M}$
\STATE Initialize skill generation $g \leftarrow 0$, RL buffer $\mathcal{B} \leftarrow \varnothing$
\FOR{each task $\tau_i$ in stream}
    \STATE \textcolor{blue}{$\triangleright$ \textit{Serve task with current meta-model}}
    \STATE $\mathcal{S}_{\tau_i} \leftarrow \textsc{Retrieve}(\mathcal{S}_g, \tau_i)$ \hfill \textcolor{gray}{\textit{// retrieve relevant skills}}
    \STATE $\xi_i \leftarrow \textsc{Execute}(\pi_{\theta}(\cdot \mid \tau_i, \mathcal{S}_{\tau_i}))$ \hfill \textcolor{gray}{\textit{// collect trajectory}}
    \STATE $r_i \leftarrow R(\xi_i)$; stamp $(\tau_i, \xi_i, r_i)$ with generation $g$
    \IF{$\xi_i$ reveals failure}
        \STATE Add $(\tau_i, \xi_i)$ to support set $\mathcal{D}^{\text{sup}}_g$
    \ELSE
        \STATE Add $(\tau_i, \xi_i, r_i, g)$ to RL buffer $\mathcal{B}$
    \ENDIF
    \STATE \textcolor{blue}{$\triangleright$ \textit{Skill-driven fast adaptation (when failures accumulate)}}
    \IF{$|\mathcal{D}^{\text{sup}}_g| \geq$ threshold}
        \STATE $\Delta\mathcal{S} \leftarrow \mathcal{E}(\mathcal{S}_g, \mathcal{D}^{\text{sup}}_g)$ \hfill \textcolor{gray}{\textit{// synthesize new skills from failures}}
        \STATE $\mathcal{S}_{g+1} \leftarrow \mathcal{S}_g \cup \Delta\mathcal{S}$ \hfill \textcolor{gray}{\textit{// evolve skill library}}
        \STATE Flush all samples with version $\leq g$ from $\mathcal{B}$ \hfill \textcolor{gray}{\textit{// support-query separation}}
        \STATE $g \leftarrow g + 1$
    \ENDIF
    \STATE \textcolor{blue}{$\triangleright$ \textit{Opportunistic policy optimization (when user is idle)}}
    \IF{OMLS detects idle window \AND $|\mathcal{B}| \geq$ batch size}
        \STATE $\theta \leftarrow \theta + \alpha \nabla_\theta \mathbb{E}_{(\tau, \xi, r, g') \sim \mathcal{B}}[R(\pi_\theta(\cdot \mid \tau, \mathcal{S}_{g'}))]$ \hfill \textcolor{gray}{\textit{// RL update}}
        \STATE Hot-swap model weights \hfill \textcolor{gray}{\textit{// deploy updated $\theta$}}
    \ENDIF
\ENDFOR
\end{algorithmic}
\end{algorithm}

\section{Experiments}
\label{sec:experiment}

\subsection{Experimental Setup}

\subsubsection{Benchmark and Evaluation Platform}

\noindent \textbf{MetaClaw-Bench.} We construct MetaClaw-Bench, a continual agentic benchmark comprising two complementary evaluation parts (934 questions total across 44 simulated workdays) for evaluating an agent's ability to adapt across a sequential stream of real-world CLI tasks. Existing agent benchmarks present tasks as independent episodes, providing no mechanism to assess whether an agent improves from accumulated experience. MetaClaw-Bench addresses this gap by structuring evaluation as multi-workday simulations in which the agent operates under consistent workspace and policy rulesets that evolve through user feedback.

1) Part~I structures evaluation as a 30-workday simulation (346 questions, days 01--30, 10--15 per day). The workspace state (files, configs, project records) persists across rounds within each day, and each question includes the evaluation outcome of the previous round as corrective feedback context. Questions fall into two types: \emph{file-check} tasks (structured edits or transformations producing output files validated by automated checkers) and \emph{multi-choice} tasks (conceptual procedural questions on domain-specific rules). Task difficulty increases monotonically with day index, with days 25--30 requiring sophisticated multi-step reasoning. Part~I's file-check tasks are heavily execution-oriented, with many interdependent side effects, providing a conservative measure of end-to-end completion.

2) Part~II extends the evaluation to a 14-workday simulation (588 questions, 42 per day: 434 multi-choice and 154 file-check). Part~II's file-check tasks are rule-based transformations where compliance with behavioral heuristics (e.g., schema conventions, timestamp formats) is the primary bottleneck, making them more amenable to skill distillation. This design provides a complementary signal: while Part~I stress-tests execution reliability, Part~II directly measures how quickly the RL-trained policy internalizes procedural rules across a higher-density task stream.

We report two primary metrics across both parts: overall accuracy (mean per-question score) and file-check completion rate (fraction of file-check outputs passing all automated checker assertions simultaneously). Because the benchmark tasks are authored to simulate realistic deployment rather than collected from actual user sessions, we view both parts as controlled stress tests of continual adaptation under increasing difficulty.

\noindent \textbf{Downstream evaluation: AutoResearchClaw.}
To test whether MetaClaw's adaptation mechanisms generalize beyond CLI-task benchmarks, we additionally evaluate on AutoResearchClaw~\citep{liu2026autoresearchclaw}, a fully autonomous 23-stage research pipeline that transforms a single research idea into a conference-ready paper, covering literature search, hypothesis generation, experiment design, code synthesis, sandbox execution, result analysis, paper drafting, and multi-agent peer review. Unlike MetaClaw-Bench's structured file-check and multi-choice tasks, AutoResearchClaw presents an open-ended, long-horizon agentic workload where failures manifest as stage retries, excessive refinement cycles, and incomplete pipeline runs. We report four pipeline-level metrics: \emph{stage retry rate}, \emph{refine cycle count}, \emph{pipeline stage completion} (out of 19 scorable stages), and a \emph{composite robustness score} (weighted average of stage completion rate at 40\%, retry reduction at 30\%, and refine cycle efficiency at 30\%).

\noindent \textbf{Baselines and Implementation Details.}
We evaluate two frontier LLMs as backbone policies: GPT-5.2~\citep{openai2025gpt52} and Kimi-K2.5~\citep{team2026kimi}. We compare three conditions: 1) Baseline: the base model served without any adaptation mechanism. 2) MetaClaw (Skills): the base model augmented with skill-driven fast adaptation; after each failed trajectory, the skill evolver synthesizes behavioral instructions immediately injected into the system prompt, with top-$k$ retrieval via cosine similarity over sentence embeddings. 3) MetaClaw (Full): the full pipeline combining skill-driven fast adaptation with opportunistic policy optimization via RL (5-day training run), evaluated for Kimi-K2.5 only, as it requires a cloud LoRA training endpoint configured for the target backbone.
All conditions use identical prompts and tool sets. This design isolates the individual contributions of the two MetaClaw components as defined in Section~\ref{sec:method}.

For the AutoResearchClaw evaluation, we deploy MetaClaw's \emph{skill-driven fast adaptation} within AutoResearchClaw's pipeline executor. After each pipeline run, failures and warnings from all 23 stages are captured as structured lessons and converted into reusable skill files via MetaClaw's lesson-to-skill evolver. On subsequent runs, accumulated skills are injected into the system prompt of all 18 LLM-driven stages. We run controlled A/B experiments with the same research topic, backbone LLM, and pipeline configuration, differing only in whether MetaClaw's skill injection is active. 

\subsection{Main Results}

Table~\ref{tab:main_results} reports performance on both parts of MetaClaw-Bench for all five model--condition pairs. MetaClaw consistently improves over the respective baselines across both models, both adaptation modes, and both benchmark parts.

\begin{table}[t]
\centering
\caption{Main results on MetaClaw-Bench Parts~I and~II. Acc.: mean per-question accuracy. Compl.: file-check completion rate. MetaClaw (Full) is evaluated for Kimi-K2.5 only. Best result per model per part is \textbf{bolded}.}
\label{tab:main_results}
\small
\renewcommand{\arraystretch}{1.3}
\begin{tabular}{@{} l l cc cc @{}}
\toprule
& & \multicolumn{2}{c}{\textbf{Part~I (30 days, 346 Q)}} & \multicolumn{2}{c}{\textbf{Part~II (14 days, 588 Q)}} \\
\cmidrule(lr){3-4}\cmidrule(lr){5-6}
\textbf{Model} & \textbf{Condition} & \textbf{Acc.\ (\%)} & \textbf{Compl.\ (\%)} & \textbf{Acc.\ (\%)} & \textbf{Compl.\ (\%)} \\
\midrule
GPT-5.2   & Baseline            & 41.1 & 14.7 & 44.9 & 58.4 \\
GPT-5.2   & MetaClaw (Skills)   & \textbf{44.0} & \textbf{17.1} & \textbf{49.1} & \textbf{67.5} \\
\midrule
Kimi-K2.5 & Baseline            & 21.4 &  2.0 & 21.1 & 18.2 \\
Kimi-K2.5 & MetaClaw (Skills)   & 28.3 &  2.0 & 26.9 & 33.8 \\
Kimi-K2.5 & MetaClaw (Full)     & \textbf{40.6} & \textbf{16.5} & \textbf{39.6} & \textbf{51.9} \\
\bottomrule
\end{tabular}
\vspace{-1em}
\end{table}

\noindent \textbf{MetaClaw improves both models and the full pipeline yields the largest gains.}
Results are consistent across both benchmark parts. For GPT-5.2, MetaClaw (Skills) raises overall accuracy from 41.1\% to 44.0\% on Part~I (+7.1\% relative) and from 44.9\% to 49.1\% on Part~II (+9.4\% relative), with file-check completion rising from 14.7\% to 17.1\% on Part~I and from 58.4\% to 67.5\% on Part~II. For Kimi-K2.5, MetaClaw (Skills) improves accuracy from 21.4\% to 28.3\% on Part~I (+32.2\%) and from 21.1\% to 26.9\% on Part~II (+27.5\%). MetaClaw (Full) yields substantially larger gains: on Part~I, accuracy reaches 40.6\% and task completion rises 8.25$\times$ (from 2.0\% to 16.5\%); on Part~II, accuracy reaches 39.6\% and file-check completion jumps from 18.2\% to 51.9\% (+185\% relative).

\noindent \textbf{Stronger models benefit less and weaker models benefit more.}
GPT-5.2 starts from a higher baseline (41.1\% vs.\ 21.4\% on Part~I), leaving less headroom for skill-driven gains. Kimi-K2.5, by contrast, lacks implicit procedural knowledge that the skill library provides explicitly, so skill injection yields larger returns. Notably, MetaClaw (Full) with Kimi-K2.5 (40.6\%) nearly closes the gap with GPT-5.2's baseline (41.1\%), demonstrating that the combination of skill injection and gradient-based policy optimization can largely compensate for model capability differences. This pattern suggests MetaClaw is particularly valuable for deploying capable but not state-of-the-art models at production scale.

\noindent \textbf{The full pipeline unlocks end-to-end task completion and skills alone do not.}
On Part~I, MetaClaw (Skills) leaves task completion rates unchanged for both models, confirming that skill injection sharpens partial execution quality without reliably enabling zero-defect outputs under heavy execution demands. MetaClaw (Full) closes this gap: Kimi-K2.5's completion rate jumps from 2.0\% to 16.5\% (8.25$\times$). On Part~II, where file-check tasks are rule-based, skills already drive a substantial completion gain (18.2\%→33.8\%), and the full pipeline pushes this further to 51.9\%, confirming that weight-level optimization provides an additive benefit on top of skill injection regardless of task type.

Since MetaClaw-Bench is an authored simulation rather than a collection of real user sessions (see Section~\ref{sec:experiment}), the absolute magnitudes of these gains are specific to this benchmark and may not transfer directly to production workloads. The primary value of these results lies in the consistent directional trends: skill-driven adaptation reliably improves partial execution quality across both models, while weight-level optimization is necessary to unlock end-to-end task completion.

\noindent \textbf{MetaClaw generalizes to open-ended multi-stage pipelines.}
Table~\ref{tab:arc_results} reports MetaClaw's impact on AutoResearchClaw, an evaluation setting structurally different from MetaClaw-Bench. Using skills-only adaptation (no RL), MetaClaw reduces the stage retry rate by 24.8\% (from 10.5\% to 7.9\%) and cuts refine cycles by 40.0\% (from 2.0 to 1.2 per stage). Pipeline completion improves from 18/19 to 19/19 stages (+5.3\%), and the composite robustness score rises from 0.714 to 0.845, an \textbf{18.3\% improvement}. These gains are achieved \emph{without any gradient-based policy updates}, demonstrating that MetaClaw's lightweight, zero-downtime skill injection transfers effectively to complex, long-horizon agentic workflows beyond structured CLI tasks.

\begin{table}[t]
\centering
\caption{MetaClaw (Skills-Only) on AutoResearchClaw, a 23-stage autonomous research pipeline. Skill injection alone yields consistent improvements across all robustness metrics without requiring RL weight updates.}
\label{tab:arc_results}
\small
\begin{tabular}{@{} l c c c @{}}
\toprule
\textbf{Metric} & \textbf{Baseline} & \textbf{+ MetaClaw (Skills)} & \textbf{Relative Change} \\
\midrule
Stage retry rate ($\downarrow$) & 10.5\% & 7.9\% & $\downarrow$ 24.8\% \\
Refine cycle count ($\downarrow$) & 2.0 & 1.2 & $\downarrow$ 40.0\% \\
Pipeline stage completion ($\uparrow$) & 18\,/\,19 & 19\,/\,19 & $\uparrow$ 5.3\% \\
Composite robustness score ($\uparrow$) & 0.714 & 0.845 & $\uparrow$ 18.3\% \\
\bottomrule
\end{tabular}
\end{table}

\subsection{Analysis}

\begin{figure}[t]
\centering
\begin{minipage}[t]{0.47\linewidth}
\centering
\includegraphics[width=\linewidth]{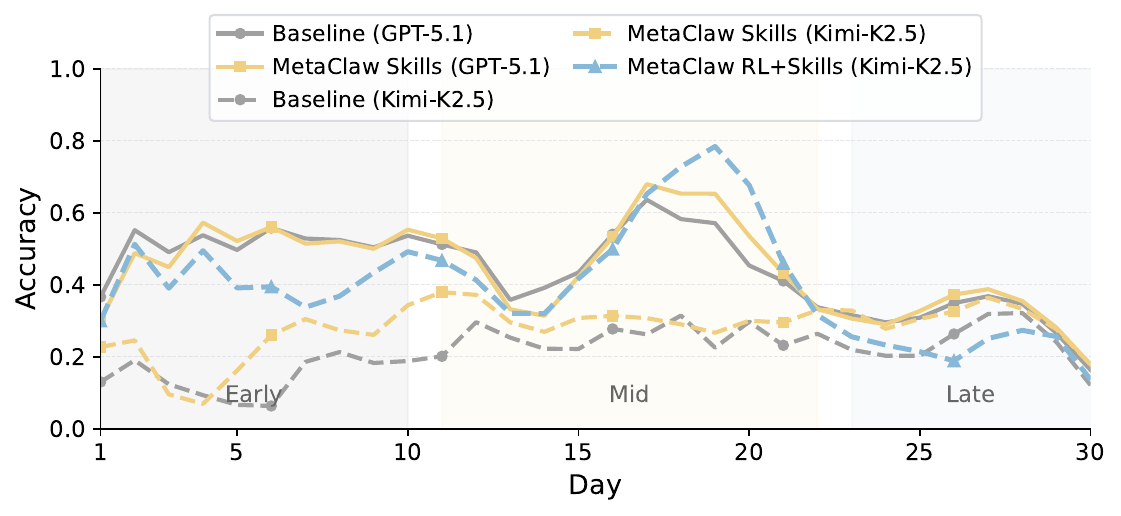}
\caption{Per-day accuracy over 30 simulated workdays (3-day rolling average). Solid lines: GPT-5.2; dashed lines: Kimi-K2.5. MetaClaw (Full) dominates in the mid phase (day 11--22) before difficulty outpaces accumulated knowledge in late days.}
\label{fig:per_day}
\end{minipage}
\hfill
\begin{minipage}[t]{0.47\linewidth}
\centering
\includegraphics[width=\linewidth]{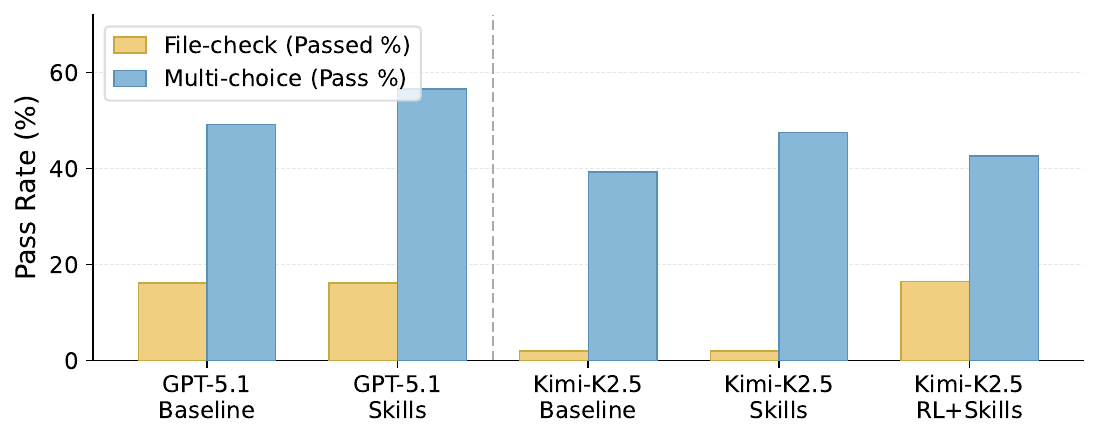}
\caption{Per-task-type pass rates. File-check (yellow) is unchanged by skills alone but jumps 8.25$\times$ under MetaClaw (Full). Multi-choice (blue) improves with skills but slightly decreases under MetaClaw (Full), reflecting a policy shift toward file-execution.}
\label{fig:task_breakdown}
\end{minipage}
\vspace{-1em}
\end{figure}

\noindent \textbf{Per-day accuracy trends.}
Figure~\ref{fig:per_day} visualizes per-day accuracy (3-day rolling average) for all five conditions. Both models and all conditions show a consistent accuracy decline from day01--10 (where accuracies routinely exceed 50\%) to day25--30 (where most models fall below 30\%), confirming that MetaClaw-Bench exhibits increasing difficulty. MetaClaw's advantage over the baseline is most pronounced in the \emph{mid-range} days (day11--22), where tasks require multi-step procedural compliance that is learnable through failure distillation, and MetaClaw (Full) reaches its peak advantage of nearly 0.8 accuracy around day~19--20. The \emph{early} days (day01--10) involve simpler manipulations where both conditions perform reasonably, and the \emph{late} days (day23--30) are sufficiently complex that accumulated skills are insufficient without stronger model weights, leading all five conditions to converge toward similarly low performance.

\noindent \textbf{Task-type breakdown.}
Figure~\ref{fig:task_breakdown} decomposes performance by task type, revealing that the two MetaClaw components address fundamentally different bottlenecks. Skills-only adaptation lifts multi-choice pass rates for both models while leaving file-check completion flat, as procedural knowledge helps reasoning but not execution. MetaClaw (Full) reverses this: Kimi-K2.5's file-check completion rate jumps to match GPT-5.2's baseline, while multi-choice accuracy slightly decreases as the policy shifts toward file-execution behavior during training.

\noindent \textbf{RL training dynamics.}
Part~II provides a fine-grained view of how policy optimization evolves over time. The file-check completion curve for MetaClaw (Full)--Kimi-K2.5 shows a clear inflection at day~8, after which per-day pass rates escalate rapidly: from $\sim$9\% on days~1--4, through 27--36\% on days~5--8, to 55--64\% on days~9--10, and ultimately reaching 100\% on days~12 and~14. This learning trajectory mirrors the MAML inner-loop update structure: the first several days accumulate support trajectories for skill synthesis and weight updates, the inflection marks when sufficient gradient signal has been collected for the LoRA fine-tune to shift the policy's execution strategy, and the late-phase convergence indicates that the policy has internalized the procedural rules surfaced by the skill library. The two-phase pattern (skill-driven gains first, RL-driven gains after day~8) directly validates the complementary timescale hypothesis underlying MetaClaw's design.

\noindent \textbf{Skill library analysis.}
Across the 30-day session, MetaClaw's skill evolver synthesizes skills clustered around three recurring failure categories: (1) \emph{temporal format compliance}, normalizing natural-language time expressions to ISO~8601 format with timezone offsets; (2) \emph{backup-before-modify protocol}, creating \texttt{.bak} files before any destructive file operation; and (3) \emph{naming convention adherence}, following date-prefixed file naming patterns (e.g., \texttt{20260408\_*.json}). These cross-cutting behavioral heuristics generalize across tasks, explaining why a single failure can yield skills that improve performance on subsequent, structurally different questions.

\noindent \textbf{Cross-domain skill transfer to AutoResearchClaw.}
The AutoResearchClaw results (Table~\ref{tab:arc_results}) provide complementary evidence for skill generalization. In this setting, the skill evolver, designed for CLI-task adaptation, synthesizes actionable skills for a fundamentally different workload (multi-stage research automation) without any domain-specific tuning. The 40\% reduction in refine cycles indicates that skills distilled from earlier pipeline failures (e.g., citation formatting errors, experiment code validation failures) directly prevent repeated mistakes in subsequent runs. This cross-domain transferability, combined with the zero-downtime deployment model (skill injection operates entirely at the prompt level), confirms that MetaClaw functions as a general-purpose continual learning layer applicable to diverse agentic systems.

\noindent \textbf{Case studies.}
Table~\ref{tab:case_study} contrasts two representative cases that illustrate the distinct contributions of the two MetaClaw mechanisms. In Case~1, a single distilled skill resolves a compliance error with zero weight update. In Case~2, skill injection provides necessary format context but is insufficient alone; weight-level RL updates are required to reliably execute a structurally complex file operation.

\begin{table}[t]
\centering
\caption{Representative case studies. Case~1 shows skill-driven fast adaptation (MetaClaw Skills, GPT-5.2); Case~2 shows the full pipeline (MetaClaw (Full), Kimi-K2.5). Both recover from score~0 to score~1.0; the mechanisms differ fundamentally.}
\label{tab:case_study}
\small
\renewcommand{\arraystretch}{1.4}
\setlength{\tabcolsep}{7pt}
\begin{tabular}{@{} l p{6.0cm} p{6.0cm} @{}}
\toprule
& \textbf{Case 1\quad MetaClaw (Skills)} & \textbf{Case 2\quad MetaClaw (Full)} \\
\cmidrule(lr){2-2}\cmidrule(lr){3-3}
Model       & GPT-5.2    & Kimi-K2.5 \\
Day / Round & Day~19 / Round~4 & Day~18 / Round~6 \\
Task type   & File-check & File-check \\
\midrule
\multicolumn{3}{@{}l@{}}{\textit{Task instruction}} \\[1pt]
& Update \texttt{sprint8\_board.json}: set T-405/T-406 to \texttt{"done"}, add \texttt{completed\_at} fields.
& Append a deployment record to \texttt{deploy\_log.json} with fields \texttt{timestamp} (ISO~8601+TZ), \texttt{env}, \texttt{status}, and \texttt{changes}. \\[3pt]
\midrule
\multicolumn{3}{@{}l@{}}{\textit{Baseline response} \textnormal{(score: 0)}} \\[1pt]
& Reads file; directly overwrites it. Checker detects missing \texttt{sprint8\_board.json.bak} $\to$ 0.
& Uses field name \texttt{date} instead of \texttt{timestamp}; omits \texttt{changes}. Checker rejects schema $\to$ 0. \\[3pt]
\midrule
\multicolumn{3}{@{}l@{}}{\textit{MetaClaw response} \textnormal{(score: 1.0)}} \\[1pt]
& Skill distilled from Day~2: \emph{``Always create \texttt{.bak} before modifying (P4).''}
  Agent writes \texttt{sprint8\_board.json.bak}, applies targeted patch. Checker passes $\to$ 1.0.
& Skills inject \emph{``use ISO~8601 with timezone offset''}; skills-only Kimi still omits \texttt{changes} array $\to$ 0.
  After RL: all four fields present, schema valid, backup created $\to$ 1.0. \\[3pt]
\midrule
\multicolumn{3}{@{}l@{}}{\textit{Day accuracy (all rounds)}} \\[1pt]
& Baseline: 43.9\% \quad MetaClaw (Skills): 62.1\% \quad \textbf{$\Delta$\,+18.2\,pp}
& Baseline: 8.3\% \quad Skills-only: 25.0\% \quad MetaClaw (Full): \textbf{80.6\%} \\[3pt]
\midrule
\multicolumn{3}{@{}l@{}}{\textit{Key mechanism}} \\[1pt]
& \textbf{Skills:} one distilled rule generalizes across file types and subsequent days with zero weight update.
& \textbf{RL:} skills supply declarative format context; weight updates internalize the execution reliability that skill injection alone cannot enforce. \\
\bottomrule
\end{tabular}
\vspace{-1.5em}
\end{table}

\section{Related Work}
\label{sec:related}

\noindent \textbf{Skill-based and memory-augmented agents.}
A line of work augments LLM agents with external memory or reusable skill libraries to improve performance without modifying model weights~\citep{shinn2023reflexion,zhao2024expel,fang2025memp,tang2025agent,ouyang2025reasoningbank,chhikara2025mem0,liu2026simplemem,wang2024agent}. Reflexion~\citep{shinn2023reflexion} stores verbal self-reflections in an episodic buffer, allowing the agent to avoid repeating past mistakes. Mem0~\citep{chhikara2025mem0} and SimpleMem~\citep{liu2026simplemem} maintain longer-horizon memory through hierarchical retrieval. On the skill side, Voyager~\citep{wangvoyager} incrementally builds a library of executable code skills from successful episodes, while ExpeL~\citep{zhao2024expel} and Agent-KB~\citep{tang2025agent} distills cross-task experience into natural-language rules. A key limitation shared by these methods is that the skill library (or memory) is treated as a static artifact~\citep{xia2026skillrl}: it is never coordinated with weight-level optimization, and successful trajectories are reused indiscriminately without regard for whether the agent's behavior has changed since they were collected. MetaClaw addresses both gaps by coupling skill evolution with policy optimization through explicit support-query separation.

\noindent \textbf{Reinforcement learning for LLM agents.}
RLHF~\citep{ouyang2022training} and its variants establish the use of reward signals to fine-tune LLM behavior, and subsequent work applies RL to tool-using and agentic settings~\citep{nakano2021webgpt,yao2022react}. More recently, GRPO~\citep{shao2024deepseekmath} and DAPO~\citep{yu2025dapo} demonstrate stable online policy gradient training for reasoning tasks~\citep{schulman2017proximal,ahmadian2024back,shao2024deepseekmath,feng2025group,zheng2025group,team2025tongyi,dong2025agentic}. However, these approaches optimize a fixed policy against a fixed reward signal, with no mechanism for the agent to update its behavioral context between rollouts. In deployed interactive settings, they also do not address \emph{when} to run training or \emph{which} data remains valid for gradient updates after behavioral changes. MetaClaw targets exactly these practical constraints via opportunistic scheduling and skill generation versioning.

\noindent \textbf{Continual and meta-learning.}
Meta-learning~\citep{finn2017model,nichol2018first,hospedales2021meta} frames learning as optimizing for fast adaptation to new tasks, typically in an offline episode-based setting. Meta-reinforcement learning extends this idea to sequential decision-making: RL\textsuperscript{2}~\citep{duan2016rl2} trains a recurrent policy whose hidden state implicitly encodes task context, PEARL~\citep{rakelly2019efficient} infers a probabilistic context variable for off-policy adaptation, and ProMP~\citep{rothfuss2019promp} applies trust-region constraints at the meta-level. These methods demonstrate effective fast adaptation in robotic control and navigation, but operate on simple network architectures with low-dimensional action spaces and assume fixed offline task distributions. Continual learning~\citep{kirkpatrick2017ewc,lopez2017gem,chaudhry2019agem,zenke2017continual,wang2024comprehensive,wang2022learning} studies sequential task adaptation without forgetting through regularization, replay, or architectural strategies, yet does not incorporate fast adaptation mechanisms at inference time. Online meta-learning approaches~\citep{finn2019online,nagabandi2018deep,harrison2020continuous,yao2020online} relax the offline assumption and even handle task heterogeneity, but remain grounded in representation learning over simple networks. MetaClaw extends the meta-learning objective to a non-stationary stream of LLM agent tasks where fast adaptation is gradient-free (skill synthesis in discrete natural-language space) and slow adaptation is gradient-based (policy optimization via RL), with a versioning protocol that preserves the support-query structure in an online, asynchronous setting.

\section{Conclusion}

We presented MetaClaw, a continual meta-learning framework that enables deployed LLM agents to improve autonomously through normal usage. MetaClaw combines two complementary adaptation mechanisms operating at different timescales: fast, inference-time skill injection that distills reusable behavioral knowledge from failures, and slow, gradient-based policy optimization that refines the model during idle windows. Built on a lightweight proxy architecture, the system requires no local GPUs and integrates transparently with existing personal agents and LLM providers. Experiments on MetaClaw-Bench demonstrate consistent improvements across models and adaptation modes, with the full pipeline yielding the largest gains on both partial execution quality and end-to-end task completion. Evaluation on AutoResearchClaw further shows that skill injection generalizes to open-ended research pipelines without any gradient updates. A current limitation is that idle-window detection depends on user configuration, which may not generalize to all deployment environments. We believe MetaClaw establishes a principled foundation for agents that genuinely learn and evolve in the wild, simply by being used.

\bibliographystyle{assets/plainnat}
\bibliography{main}

\clearpage
\newpage
\beginappendix

\section{Prompts and Templates}
\label{appendix:prompts}

This appendix documents the core prompt templates used in MetaClaw-Bench evaluations and the MetaClaw framework components.

\subsection{Agent System Prompt (MetaClaw-Bench Part~I)}
\label{appendix:system_prompt_p1}

Part~I evaluates agents on OpenClaw CLI tasks via a programmatic rollout loop. The agent receives the following fixed system prompt, which may be replaced by a compressed variant (see \cref{appendix:compression_prompt}) after the first session:

\begin{tcolorbox}[
  colback=gray!5, colframe=gray!40,
  title={\small\texttt{SYSTEM\_PROMPT} — Part~I Agent},
  fonttitle=\bfseries\small, left=6pt, right=6pt, top=4pt, bottom=4pt
]
\small
\begin{verbatim}
You are an expert CLI agent controlling an OpenClaw installation.
Your goal is to complete the given task by issuing CLI commands
via the run_command tool.

Guidelines:
- Issue ONE command at a time and carefully read the output
  before proceeding.
- Use 'openclaw status' or similar read commands to inspect
  state before making changes.
- Handle errors: if a command fails, diagnose from the output
  and retry differently.
- When the task is fully complete, call run_command with
  command="done".
- Do NOT ask clarifying questions — act based on the task
  description alone.
\end{verbatim}
\end{tcolorbox}

\noindent The single tool exposed to the agent is \texttt{run\_command}:

\begin{tcolorbox}[
  colback=gray!5, colframe=gray!40,
  title={\small\texttt{run\_command} Tool Schema},
  fonttitle=\bfseries\small, left=6pt, right=6pt, top=4pt, bottom=4pt
]
\small
\begin{verbatim}
{
  "name": "run_command",
  "description": "Execute a CLI command and observe the output.
    When the task is fully complete, call run_command with
    command=\"done\".",
  "parameters": {
    "type": "object",
    "properties": {
      "command": {
        "type": "string",
        "description": "The exact CLI command to run, e.g.
          'openclaw status',
          'openclaw agents add --name bot1 --model gpt-4o',
          'done'"
      }
    },
    "required": ["command"]
  }
}
\end{verbatim}
\end{tcolorbox}

\subsection{Agent Identity Context (MetaClaw-Bench Part~II)}
\label{appendix:identity_p2}

Part~II (CALMB-14) injects the following workspace context files into the agent's session at initialization, defining the agent's role, user profile, and behavioral principles.

\begin{tcolorbox}[
  colback=gray!5, colframe=gray!40,
  title={\small\texttt{IDENTITY.md} — Agent Role Definition},
  fonttitle=\bfseries\small, left=6pt, right=6pt, top=4pt, bottom=4pt
]
\small
\begin{verbatim}
# Identity
You are MetaClaw Agent, an AI assistant integrated into
the internal toolchain of Orion Tech.

## Role
You serve as Alex Zhang's primary AI assistant for day-to-day
engineering and project management work.

## Context
- Company:   Orion Tech — a B2B SaaS company
- Product:   Project Orion, a project management platform
- Team:      Backend engineering team
- Principal: Alex Zhang, Backend Tech Lead

## Principles
1. Accuracy comes first — double-check facts, values, and
   formats before writing files
2. Be consistent — use the same conventions across all files
3. Be complete — fill in all required fields; no placeholders
4. Be professional — produce output ready for team use
\end{verbatim}
\end{tcolorbox}

\begin{tcolorbox}[
  colback=gray!5, colframe=gray!40,
  title={\small\texttt{SOUL.md} — Core Behavioral Principles},
  fonttitle=\bfseries\small, left=6pt, right=6pt, top=4pt, bottom=4pt
]
\small
\begin{verbatim}
## Reliability
Every file you create or modify should be correct and complete.
A half-finished or incorrect file creates more work than it saves.

## Consistency
When you establish a pattern in one file, maintain it across
all related files.

## Attention to Detail
Small errors in data files can cascade into larger problems.
Pay close attention to field names, data types, value formats,
and structural requirements.

## Ownership
Own every assigned task completely. Do not produce output
requiring the user to fix obvious issues.

## Professionalism
All output should meet the standard of work that could be
shared directly with teammates or stakeholders.
\end{verbatim}
\end{tcolorbox}

\subsection{Task Question Templates}
\label{appendix:question_templates}

Each day in MetaClaw-Bench presents two question types. Below are representative examples.

\noindent\textbf{Multi-choice question (Part~II, Day~01 / r1):}

\begin{tcolorbox}[
  colback=blue!3, colframe=blue!25,
  title={\small Multi-Choice Question Template},
  fonttitle=\bfseries\small, left=6pt, right=6pt, top=4pt, bottom=4pt
]
\small
\begin{verbatim}
Regarding the source and applicability of the baseline daily
revenue of 4500 yuan/store in the financial model, which of
the following descriptions are consistent with the project
documentation? (Select all correct options)

A. The assumption document labels 4500 yuan as "Source:
   operational baseline data from market research report",
   while the original description in the market research
   report is "median of 86 tier-1 city stores"
B. The rent assumption of 650 yuan/m2 is noted as tier-2 and
   tier-3 city market rates, but the revenue assumption of
   4500 yuan references tier-1 city store data
C. Operations VP Zhang Wei mentioned an estimated daily revenue
   of approximately 2500-3200 yuan, but this is only a
   hypothetical opinion with no direct contradiction
D. The validation memo confirms that 4500 yuan "has been
   cross-validated with operational baseline data", indicating
   the assumption underwent applicability review

Please answer using \bbox{X} or \bbox{X,Y} format.
\end{verbatim}
\tcblower
\small \textbf{Ground truth:} A, B \quad\quad \textbf{Scoring:} $\max(0,\; 1 - (FP + FN) / n\_\text{options})$
\end{tcolorbox}

\noindent\textbf{File-check question (Part~II, Day~01 / r21):}

\begin{tcolorbox}[
  colback=blue!3, colframe=blue!25,
  title={\small File-Check Question Template},
  fonttitle=\bfseries\small, left=6pt, right=6pt, top=4pt, bottom=4pt
]
\small
\begin{verbatim}
Based on the reference documents in day01/, create a decision
log tracking key decisions from the documents. Save as
day01/decision_log_r21.json. Include these fields: title,
created_at, decisions (array with id, date, decision,
rationale, decided_by, review_date).
\end{verbatim}
\tcblower
\small \textbf{Eval:} \texttt{python scripts/check\_iso8601.py day01/decision\_log\_r21.json} (exit~0~=~pass) \\
\textbf{Feedback (incorrect):} \textit{Time/date fields must use ISO 8601 with +08:00 timezone: YYYY-MM-DDTHH:MM:SS+08:00.}
\end{tcolorbox}

\noindent\textbf{Part~I task instruction format} (real OpenClaw session, \texttt{train.jsonl}):

\begin{tcolorbox}[
  colback=blue!3, colframe=blue!25,
  title={\small Part~I User Instruction Template},
  fonttitle=\bfseries\small, left=6pt, right=6pt, top=4pt, bottom=4pt
]
\small
\begin{verbatim}
[Sat 2026-02-21 07:25 EST] I grant you read access to
/Users/jimchen/Documents/openclaw/skills. Locate gog/skill.md
within it. Use skill.md to obtain the instructions for adding
a Google Calendar. Execute the instructions in the isolated
environment to directly add ten meetings to Google Calendar,
starting February 20th, recurring every Friday from
3:30-4:30 PM.

Language requirement: Reply in English only. Keep your
response concise and task-focused.
\end{verbatim}
\end{tcolorbox}

\subsection{Skill Evolver Prompt Template}
\label{appendix:skill_evolver_prompt}

After each session in which failed trajectories are collected, the MetaClaw skill evolver submits the following prompt to synthesize new behavioral skills:

\begin{tcolorbox}[
  colback=green!4, colframe=green!30,
  title={\small Skill Evolver Prompt (\texttt{skill\_evolver.py})},
  fonttitle=\bfseries\small, left=6pt, right=6pt, top=4pt, bottom=4pt
]
\small
\begin{verbatim}
You are a skill engineer for an AI assistant trained with RL.
Your job: analyze the failed conversations below and generate
NEW skills that would have prevented those failures.

---
## Failed Conversations

### Failure 1  (reward=0.0)
**Conversation context (last 600 chars):**
```
...[truncated trajectory]
```
**Assistant response (first 500 chars):**
```
[model response]
```
[...up to 6 failures shown...]

---
## Existing Skills (do NOT duplicate any of these)
["skill-name-1", "[category] skill-name-2", ...]

---
## Instructions

Generate **1 to {max_new_skills}** new skills that directly
address the failure patterns observed above. Focus on
actionable, concrete guidance for future conversations.

Each skill must follow Claude skill format:
- `name`: a lowercase hyphenated slug
- `description`: one sentence — when to trigger this skill
  and what it achieves
- `content`: 6-15 lines of actionable Markdown. Include:
  a heading, numbered steps or bullet points, a concrete
  example or code snippet, and an Anti-pattern section.
- `category`: one of [coding, research, data_analysis,
  security, communication, automation, productivity, agentic]
  or "general" or "common_mistakes"

**Output:** Return ONLY a valid JSON array.

**Example output:**
[
  {
    "name": "dyn-001",
    "description": "Always verify file existence before
      reading or writing.",
    "content": "## Verify File Existence Before Acting\n\n
      1. Check: os.path.exists(path)\n
      2. If missing, ask the user for the correct path.\n
      **Anti-pattern:** Calling open(path) without checking.",
    "category": "coding"
  }
]
\end{verbatim}
\end{tcolorbox}

\subsection{Skill Injection Format}
\label{appendix:skill_injection}

Retrieved skills are appended to the agent's system message by \texttt{SkillManager.format\_for\_conversation}:

\begin{tcolorbox}[
  colback=green!4, colframe=green!30,
  title={\small Skill Injection Block (appended to system prompt)},
  fonttitle=\bfseries\small, left=6pt, right=6pt, top=4pt, bottom=4pt
]
\small
\begin{verbatim}
## Active Skills

### backup-before-modify
_Always create a .bak copy before modifying any existing file._

## Backup Before Modify

1. Before editing any file, create a backup:
   cp <filename> <filename>.bak
2. Verify the backup exists before proceeding.
3. Apply all modifications to the original file.

**Anti-pattern:** Overwriting a file without a backup,
leaving no recovery path if the edit is incorrect.

### iso8601-timezone-format
_Use when writing any date/time field to a file._

## ISO 8601 Timestamp with Timezone

Always format timestamps as: YYYY-MM-DDTHH:MM:SS+08:00

- Correct:   2026-03-16T09:30:00+08:00
- Incorrect: 2026-03-16, March 16 at 3pm,
             2026-03-16T09:30:00Z

**Anti-pattern:** Omitting the timezone offset or using
natural-language date expressions.
\end{verbatim}
\end{tcolorbox}

\subsection{System Prompt Compression Prompt}
\label{appendix:compression_prompt}

To prevent context overflow during long sessions, MetaClaw compresses OpenClaw's native system prompt using the following instruction:

\begin{tcolorbox}[
  colback=gray!5, colframe=gray!40,
  title={\small System Prompt Compression Instruction (\texttt{utils.py})},
  fonttitle=\bfseries\small, left=6pt, right=6pt, top=4pt, bottom=4pt
]
\small
\begin{verbatim}
You are compressing an OpenClaw system prompt. Rewrite it to
be under 2000 tokens while preserving behavior. Keep all
critical policy and routing rules:
(1) tool names and their intended usage constraints,
(2) safety and non-delegable prohibitions,
(3) skills-selection rules,
(4) memory recall requirements,
(5) update/config restrictions,
(6) reply-tag/messaging rules,
(7) heartbeat handling rules.
Remove duplicated prose, repeated examples, and decorative
language. Prefer compact bullet sections with short imperative
statements. Do not invent or weaken any rule. Output only
the rewritten system prompt text.
\end{verbatim}
\end{tcolorbox}

\subsection{Part~II Implicit Preference Rules}
\label{appendix:preference_rules}

MetaClaw-Bench Part~II introduces five implicit preference rules progressively across 14 days. These rules are \emph{not} stated in the agent's system prompt; they must be inferred from task feedback and internalized through skill evolution or RL training.

\begin{table}[H]
\centering
\small
\renewcommand{\arraystretch}{1.35}
\begin{tabular}{@{} l l p{6.5cm} l @{}}
\toprule
\textbf{Rule} & \textbf{Category} & \textbf{Requirement} & \textbf{Active from} \\
\midrule
P1 & Timestamp  & All date/time fields: \texttt{YYYY-MM-DDTHH:MM:SS+08:00} & Day 01 \\
P2 & File naming & Output files: \texttt{YYYYMMDD\_description.ext} (snake\_case) & Day 04 \\
P3 & Metadata   & Every output file must include \texttt{created\_at}, \texttt{author}, \texttt{status} & Day 06 \\
P4 & Backup     & Create \texttt{<file>.bak} before modifying any existing file & Day 08 \\
P5 & Completion log & Append \texttt{[DONE] <timestamp> | <task\_id> | <summary>} to \texttt{done.log} & Day 10 \\
\bottomrule
\end{tabular}
\caption{The five implicit preference rules in MetaClaw-Bench Part~II, introduced progressively across 6 learning arcs. Each rule is verified by a dedicated automated checker script.}
\label{tab:preference_rules}
\end{table}

\end{document}